\documentclass{article}

    \PassOptionsToPackage{numbers, compress}{natbib}

\usepackage[preprint]{neurips_2026}

\usepackage{float}
\usepackage{wrapfig}
\usepackage{graphicx}
\usepackage[utf8]{inputenc} 
\usepackage[T1]{fontenc}    
\usepackage{hyperref}       
\usepackage{url}            
\usepackage{booktabs}       
\usepackage{amsfonts}       
\usepackage{nicefrac}       
\usepackage{microtype}      
\usepackage{xcolor}         
\usepackage{amsmath}
\usepackage{multirow}

\title{Towards Localized and Disentangled Knowledge Editing for Multimodal Large Language Models}

\author{%
  Leijiang Gu\textsuperscript{1}, 
  Zhen Zeng\textsuperscript{1}, 
  Feng Li\textsuperscript{1}, 
  Xinjian Gao\textsuperscript{1}, 
  Xiaofeng Cao\textsuperscript{2}, 
  Zenglin Shi\textsuperscript{1}\thanks{Corresponding author.} \\[0.2cm] 
  \textsuperscript{1}Hefei University of Technology, Hefei, China \\
  \textsuperscript{2}Tongji University, Shanghai, China \\[0.2cm] 
  \texttt{2024170839@mail.hfut.edu.cn, zenglin.shi@hfut.edu.cn}
}

\begin{document}

\maketitle

\begin{abstract}
Existing methods in Multimodal Knowledge Editing (MKE) have advanced the ability to correct outdated or inaccurate knowledge in Multimodal Large Language Models (MLLMs). However, they exhibit a critical limitation: while effectively modifying target factual pairs, they fail to generalize edits to logically related queries and often cause unintended alterations to unrelated but visually or semantically linked information. We identify and formalize two underlying failure modes causing this issue: Causal Misalignment, which confines edits to the specific sample, and Feature Entanglement, which causes unintended alterations to coupled but irrelevant information. To address these issues, we propose Localized and Disentangled Knowledge Editing (LDKE), a new framework that achieves precise and generalized editing by localizing fact-specific model layers and disentangling target-relevant inputs from irrelevant ones. Our approach introduces a Fast Localization module to identify and update critical layers efficiently, along with a Disentanglement Classifier that routes inputs appropriately to preserve unrelated knowledge. Extensive experiments across various benchmarks and MLLMs demonstrate that LDKE achieves superior performance in propagating edits to related contexts while maintaining high locality.

\end{abstract}

\section{Introduction}

\begin{figure}
  \centering
  \includegraphics[width=1.0\linewidth]{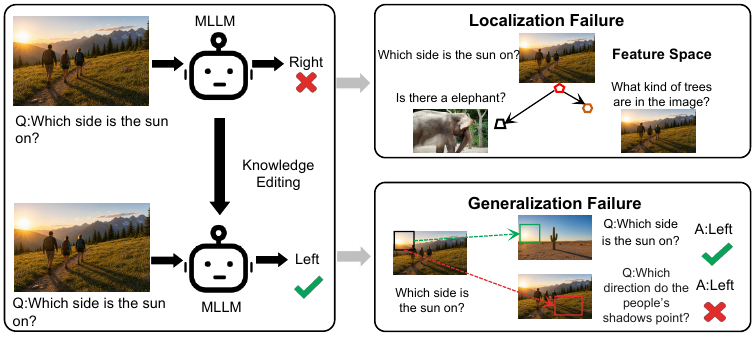}
  \caption{Illustration of the generalization-localization challenge in multimodal knowledge editing. Existing methods can correct the edited query, but the update may not transfer to logically related queries and may disturb unrelated visual-semantic content.}
  \label{fig:main_pic}
\end{figure}

Multimodal large language models~(MLLMs) have achieved remarkable success in integrating visual and textual information~\cite{yang2023mm,alayrac2022flamingo}, driving their deployment across various applications~\cite{llama2023llama,mishra2024fine,ishibashi2023knowledge,yin2024woodpecker}. As the knowledge encoded in MLLMs can become outdated or inaccurate over time~\cite{knowledgeable,zhang2024comprehensive}, multimodal knowledge editing~(MKE) has emerged as an efficient way to update specific visual-semantic facts without full retraining~\cite{knowledgeediting,hartvigsen2023aging,visedit}. However, effective MKE requires more than correcting the response to a single edited image-question pair: the updated knowledge should reliably generalize to logically related queries while remaining well localized to the intended visual-semantic fact. Existing methods still struggle to balance these two requirements. Although they can often fit the edited sample and its simple rephrases, their updates may not transfer sufficiently to logically related queries and can inadvertently affect unrelated knowledge that is visually or semantically coupled with the edited fact. As illustrated in Fig.~\ref{fig:main_pic}, this reveals a central generalization-localization challenge in multimodal knowledge editing.

We attribute the difficulty of this generalization-localization challenge to two underlying factors: causal misalignment and feature entanglement. Causal misalignment refers to the mismatch between the layers modified by editing and the causal modules that mediate factual associations. Many existing methods rely on fixed-layer updates that primarily fit the target output, which can correct the original edit prompt but does not necessarily modify the internal association required for logical transfer. Although causal tracing has shown that intermediate feed-forward networks play a key role in factual association recall and has been extended to MLLMs~\citep{causal_hallucination,causal_information}, its high computational cost makes instance-level dynamic localization impractical, forcing methods to rely on static localization and limiting generalization. Feature entanglement arises when the target fact shares highly similar representations with unrelated visual-semantic content. Under this condition, preservation constraints can conflict with the editing objective, causing updates to affect feature-proximal but irrelevant knowledge. Similarly, standard classifiers only impose surface decision boundaries over entangled representations; when relevant and irrelevant facts are densely coupled, these boundaries become fragile and prone to routing unrelated queries to edited weights.

To address the dual challenges of causal misalignment and feature entanglement, we propose the Localized and Disentangled Knowledge Editing~(LDKE) framework. 
For causal misalignment, LDKE replaces fixed-layer editing with an instance-aware Fast Localization module. 
Instead of relying on traditional causal tracing, which is prohibitively expensive for MLLMs, Fast Localization estimates the contribution of each FFN layer by measuring the target probability delta before and after the FFN transformation. 
This allows LDKE to identify the fact-critical layers for each input in a single forward pass, avoiding output-proximal fitting and enabling more targeted updates to the underlying factual association. 
The localized layers then guide a Layer-Specific Weight Editor to generate precise weight updates. For feature entanglement, LDKE introduces a Disentanglement Classifier to prevent edited weights from affecting unrelated knowledge. The classifier decouples the hidden representation of the input and compares it with the edit representation through cosine similarity. This similarity score serves as a dynamic routing gate: inputs within the edit scope are routed through the edited weights, while out-of-scope inputs are processed by the original frozen weights. By combining instance-aware localization with representation-level disentanglement, LDKE aims to update related knowledge consistently while minimizing unintended interference with unrelated knowledge.
To rigorously validate our approach, we. conduct extensive evaluations on the multiple benchmarks, achiving better performance.

The main contributions of this work are summarized as follows:

\begin{itemize}
    \item  We systematically identify the critical bottlenecks of current multimodal knowledge editing, namely causal misalignment and feature entanglement.
    \item  We introduce LDKE framework, composed of Fast Localization module and  Disentanglement Classifier. Fast Localization module dynamically pinpoints fact-critical layers for individual samples via a single forward pass. Disentanglement Classifier decouples hidden state and leverages similarity scores to isolate unrelated queries.
    \item Extensive experiments on the multiple benchmarks demonstrate that the LDKE framework significantly outperforms existing approaches, validating its efficacy in overcoming the intricate challenges of causal misalignment and feature entanglement.
\end{itemize}

\section{Related work}

\subsection{Knowledge Editing for LLMs}
\noindent Existing knowledge editing methods for LLMs can be broadly divided into parameter-preserving methods and parameter-modifying methods. \textbf{Parameter-preserving methods} edit the model through external memory or additional weights without directly changing the original parameters. SERAC~\citep{serac} augments the base model with an explicit memory to store edited facts and a classifier to determine whether a query falls within the edited scope. T-Patcher~\citep{t-patch} and CaliNET~\citep{calibrating} add extra neurons to FFNs and train them to achieve model editing. IKE~\citep{IKE} utilizes demonstration prompts to inject new facts without parameter optimization. \textbf{Parameter-modifying methods} directly revise model weights to update factual associations. MEND~\citep{MEND} trains small auxiliary editing networks to transform the gradient obtained by standard fine-tuning into weight updates by a low-rank decomposition of the gradient. ROME~\citep{rome} uses causal tracing to reveal that middle-layer feed-forward modules are responsible for processing fact associations. MEMIT~\citep{MEMIT} extends the ROME framework to accommodate batch editing, enhancing its efficacy in handling simultaneous factual modifications. 

\subsection{Knowledge Editing for MLLMs}
Currently, research dedicated to multimodal knowledge editing remains relatively scarce. Building upon the SERAC framework, MSCKE~\citep{mscke} extends the scope classifier to integrate both visual and textual modalities. Through contribution allocation and noise perturbation, VisEdit~\citep{visedit} identifies that prompt-relevant visual representations in mid-to-later layers heavily influence factual predictions and modifies these specific intermediate visual related to edit prompt. To address lifelong knowledge editing, LiveEdit~\citep{liveedit} dynamically generates instance-specific low-rank experts and employs a two-stage visual-semantic routing mechanism to efficiently activate relevant experts during inference.
Despite these advances, existing knowledge editing methods still struggle with the generalization-localization challenge, which motivate the LDKE framework.


\section{Method}
\begin{figure}[t] 
    \centering
    \includegraphics[width=\textwidth]{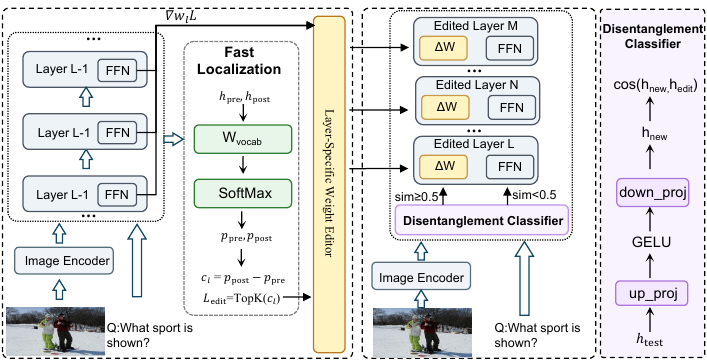} 
    
    \caption{The overall framework of LDKE. LDKE consists of a Fast Localization module and a Disentanglement Classifier.}
    \label{fig:method}
\end{figure}

In this section, we present our Locate and Distanglement (LDKE) framework, the overall architecture of which is illustrated in Fig.~\ref{fig:method}. LDKE consists of two core components, a Fast Localization module and a Disentanglement Classifier. The Fast Localization module dynamically identifies the critical FFN layers responsible for fact association via a single forward pass per instance. Then, we utilize the Layer-Specific Weight Editor to generate update weights for target layers. Meanwhile, the Disentanglement Classifier decouples the hidden representation of the last prompt token and computes its cosine similarity against the target edit representation, effectively determining whether to apply the updated weights.

\subsection{Problem Formulation}
Consider a MLLM $f_{\theta}$ parameterized by $\theta$. For a specific fact, given an edit image $i_e$ and an edit textual prompt $t_e$, the original model generates an output $y_o = f_{\theta}(i_e, t_e)$, where $y_o$ is typically the incorrect or outdated information.
The goal of multimodal knowledge editing is to define a knowledge update operation $\mathcal{U}$ that modifies the original parameters $\theta$ to yield $\theta_e = \mathcal{U}(\theta, i_e, t_e, y_e)$, where $y_e$ is the desired target correction. Following this intervention, the edited model $f_{\theta_e}$ must reliably generate the target output for the edit prompt, satisfying $f_{\theta_e}(i_e, t_e) = y_e$.

\subsection{Instance-Aware Fast Localization}
\label{Fast Localization}

Existing editing methods such as ROME~\citep{rome} identify editable layers through causal tracing, which helps locate factual associations but requires repeated forward interventions. This cost becomes prohibitive for MLLMs, where dense visual tokens significantly increase sequence length. To enable efficient instance-level localization, we introduce a Fast Localization module that estimates each FFN layer's factual contribution in a single forward pass.

Prior causal-tracing analysis of object representations in LVLMs shows that middle-layer last-token MHSAs are critical for cross-modal aggregation and that FFNs exhibit a hierarchical progression in storing and transferring visual object representations~\citep{causal_hallucination}. Based on this finding, we use the hidden representation of the last prompt token as the localization signal. Given an edit instance $(i_e,t_e,y_e)$, we run the original model $f_\theta$ once and collect the hidden representation of the last prompt token before and after each FFN block. Let $\mathbf{h}^{(\ell)}_{\text{pre}}$ and $\mathbf{h}^{(\ell)}_{\text{post}}$ denote the last-token representations immediately before and after the FFN module at layer $\ell$, respectively. We project both representations into the vocabulary space with the unembedding matrix $\boldsymbol{W}_{\text{vocab}}$ and compute the probability assigned to the target answer:
\begin{equation}
P^{(\ell)}_{\text{pre}}(y_e)
=
\left[
\mathrm{Softmax}
\left(
\boldsymbol{W}_{\text{vocab}}\mathrm{LN}(\mathbf{h}^{(\ell)}_{\text{pre}})
\right)
\right]_{y_e},
\end{equation}

\begin{equation}
P^{(\ell)}_{\text{post}}(y_e)
=
\left[
\mathrm{Softmax}
\left(
\boldsymbol{W}_{\text{vocab}}\mathrm{LN}(\mathbf{h}^{(\ell)}_{\text{post}})
\right)
\right]_{y_e},
\end{equation}

where $\mathrm{LN}(\cdot)$ denotes Layer Normalization.
For a more stable contribution estimate, we estimate the contribution of the $\ell$-th FFN module by its induced log-probability shift:
\begin{equation}
c_\ell
=
\log P^{(\ell)}_{\text{post}}(y_e)
-
\log P^{(\ell)}_{\text{pre}}(y_e).
\end{equation}
A larger positive value of $c_\ell$ indicates that the corresponding FFN module more strongly promotes the desired target answer. Following prior observations~\citep{causal_hallucination} that causal effects in MLLMs are mainly concentrated in middle-to-late layers, we restrict the candidate set to the latter half of the network:
\begin{equation}
\mathcal{C}
=
\{\ell \mid \ell \ge \lfloor N/2 \rfloor\},
\end{equation}
where $N$ is the total number of Transformer layers and $k$ is the number of edited layers. The target editing layers are then selected as:
\begin{equation}
\mathcal{L}_{\text{edit}}
=
\mathrm{TopK}_{\ell\in\mathcal{C}}(c_\ell,k).
\end{equation}
Here, $\mathrm{TopK}_{\ell\in\mathcal{C}}(c_\ell,k)$ returns the $k$ layers in $\mathcal{C}$ with the largest contribution scores.

\subsection{Layer-Specific Weight Editor}

To translate raw gradients into precise, layer-specific modifications, we train a hypernetwork-based weight editor $\Phi_\phi$ parameterized by $\phi$ and adapted from the MEND architecture~\cite{MEND}. During the training phase, given an edit instance $(i_e, t_e, y_e)$, we first compute the standard auto-regressive loss $\mathcal{L}_{\text{ar}}$ to derive the raw weight gradient $\nabla_{\boldsymbol{W}^{(\ell)}} \mathcal{L}_{\text{ar}} = \boldsymbol{\delta}_\ell \mathbf{x}_\ell^T$ for a designated target layer $\ell$, where $\mathbf{x}_\ell$ is the input activation and $\boldsymbol{\delta}_\ell$ is the output gradient.
To minimize parameter overhead, $\Phi_\phi$ shares low-rank projection matrices ($\boldsymbol{A}_i, \boldsymbol{B}_i$) across layers with identical shapes, which are then modulated by stage-specific layer scales ($\boldsymbol{\gamma}_\ell^{(i)}$) and offsets ($\boldsymbol{\beta}_\ell^{(i)}$), where $i\in\{1,2\}$. Taking the concatenated activation and output gradient $\mathbf{z}_\ell = [\mathbf{x}_\ell; \boldsymbol{\delta}_\ell]$ as input, the editor network computes the update representation:
\begin{align}
    \mathbf{m}_\ell &= \mathbf{z}_\ell + \sigma \left( \boldsymbol{\gamma}_\ell^{(1)} \odot (\boldsymbol{A}_1 \boldsymbol{B}_1 \mathbf{z}_\ell + \mathbf{b}) + \boldsymbol{\beta}_\ell^{(1)} \right), \\
    \Phi_\phi(\mathbf{z}_\ell) &= \mathbf{m}_\ell + \sigma \left( \boldsymbol{\gamma}_\ell^{(2)} \odot (\boldsymbol{A}_2 \boldsymbol{B}_2 \mathbf{m}_\ell) + \boldsymbol{\beta}_\ell^{(2)} \right),
\end{align}
where $\sigma(\cdot)$ denotes the activation function and $\mathbf{b}$ is a bias vector. The output $\Phi_\phi(\mathbf{z}_\ell)$ is subsequently partitioned to formulate the final target weight perturbation $\Delta \boldsymbol{W}^{(\ell)}$. Notably, for advanced architectures with gated MLPs (e.g., Gemma-3, InternVL-3.5), we strictly confine this editing scope to the core up- and down-projection matrices, omitting the gating projections.

\subsection{Disentanglement-Based Routing}
Following the generation of layer-wise weight perturbations $\{\Delta \boldsymbol{W}^{(\ell)}\}_{\ell\in\mathcal{L}_{\text{edit}}}$, strictly maintaining locality to avoid collateral disruptions to feature-proximal yet unrelated knowledge remains a paramount challenge. To solve this problem, we propose a novel Disentanglement Classifier functioning as a dynamic routing gate, effectively isolating edited weights from unrelated facts.

Given a test query $q_{\text{test}}=(i_{\text{test}}, t_{\text{test}})$, we perform routing before the forward pass reaches any edited layer. Let $\ell_{\min}=\min(\mathcal{L}_{\text{edit}})$ denote the earliest edited layer, and extract the last-token hidden state $\mathbf{h}^{(\ell_{\min})}_{\text{test}}$ at this layer. The Disentanglement Classifier then maps $\mathbf{h}^{(\ell_{\min})}_{\text{test}}$ into an intermediate router representation $\mathbf{r}_{\text{test}}$ through a lightweight residual projection, and feeds $\mathbf{r}_{\text{test}}$ into two parallel heads: an embedding head for similarity-based routing and a classification head for binary supervision. The embedding head produces an $L_2$-normalized representation, while the classification head outputs a scalar logit for the BCE loss.

\begin{equation}
\mathbf{r}_{\text{test}} = \mathbf{h}^{(\ell_{\min})}_{\text{test}} + \boldsymbol{W}_{\text{down}}\sigma(\boldsymbol{W}_{\text{up}}\mathrm{LN}(\mathbf{h}^{(\ell_{\min})}_{\text{test}})),
\label{eq:router_hidden}
\end{equation}
\begin{equation}
\tilde{\mathbf{h}}^{(\ell_{\min})}_{\text{test}} =
\frac{\boldsymbol{W}_{\text{emb}}\mathbf{r}_{\text{test}}}{\|\boldsymbol{W}_{\text{emb}}\mathbf{r}_{\text{test}}\|_2},
\label{eq:router_embedding}
\end{equation}
where $\boldsymbol{W}_{\text{up}}$, $\boldsymbol{W}_{\text{down}}$, and $\boldsymbol{W}_{\text{emb}}$ are router projection matrices.
Similarly, we compute the anchored edit representation $\tilde{\mathbf{h}}^{(\ell_{\min})}_e$ by applying Eqs.~\eqref{eq:router_hidden} and \eqref{eq:router_embedding} to the edit instance.

To determine whether this query belongs to the relevant fact or the unrelated locality fact, we compute its cosine similarity against $\tilde{\mathbf{h}}^{(\ell_{\min})}_e$:

\begin{equation}s_{\text{test}} = \frac{\tilde{\mathbf{h}}^{(\ell_{\min})}_{\text{test}} \cdot \tilde{\mathbf{h}}^{(\ell_{\min})}_e}{\|\tilde{\mathbf{h}}^{(\ell_{\min})}_{\text{test}}\|_2 \|\tilde{\mathbf{h}}^{(\ell_{\min})}_e\|_2} = \cos(\tilde{\mathbf{h}}^{(\ell_{\min})}_{\text{test}}, \tilde{\mathbf{h}}^{(\ell_{\min})}_e)\label{eq:inference_sim}\end{equation}

Thanks to the absolute similarity regularization enforced during training, the feature space presents a clear geometric boundary. We employ an absolute similarity threshold of 0.5 to explicitly categorize the incoming query. The dynamic routing gate $g \in \{0, 1\}$ is formally defined by the indicator function:

\begin{equation}
g = \begin{cases}
1, & s_{\text{test}} \geq 0.5 \\
0, & s_{\text{test}} < 0.5
\end{cases}
\end{equation}

Ultimately, the final weights $\boldsymbol{W}_{\text{final}}^{(\ell)}$ used for the forward pass at any targeted layer $\ell \in \mathcal{L}_{\text{edit}}$ are dynamically composed as:

\begin{equation}\boldsymbol{W}_{\text{final}}^{(\ell)} = g \cdot (\boldsymbol{W}_{\text{original}}^{(\ell)} + \Delta \boldsymbol{W}^{(\ell)}) + (1 - g) \cdot \boldsymbol{W}_{\text{original}}^{(\ell)} \end{equation}

\subsection{Training of LDKE} The training of LDKE involves two learnable modules: the Layer-Specific Weight Editor and the Disentanglement Classifier. The Fast Localization module is training-free and is only activated during inference to select instance-specific edited layers. Accordingly, we optimize the editor and the classifier with two losses, respectively targeting reliable weight generation and disentangled routing.

\paragraph{Editor Loss.} The Layer-Specific Weight Editor is trained to generate precise weight perturbations that correct the target knowledge while preserving unrelated information. Following the standard knowledge editing setting, we optimize the edited model with both generality and locality constraints. The generality loss encourages the edited model to produce the desired answer for rephrased queries, while the locality loss penalizes undesired changes on unrelated inputs. To handle multimodal scenarios, we further introduce multimodal generality and multimodal locality loss. The editor loss is formulated as:
\begin{equation}
\mathcal{L}_{\text{editor}}
=
\lambda_{\text{gen}}\mathcal{L}_{\text{gen}}
+
\lambda_{\text{loc}}\mathcal{L}_{\text{loc}}
+
\lambda_{\text{m\_gen}}\mathcal{L}_{\text{m\_gen}}
+
\lambda_{\text{m\_loc}}\mathcal{L}_{\text{m\_loc}},
\end{equation}
where $\lambda_{\text{gen}}$, $\lambda_{\text{loc}}$, $\lambda_{\text{m\_gen}}$, and $\lambda_{\text{m\_loc}}$ are balancing coefficients.
\paragraph{Disentanglement Loss.} The Disentanglement Classifier is trained to construct a routing space in which edit-related queries are separated from feature-proximal yet unrelated queries. Applying the same router embedding to the edit sample, a logically related query, and a feature-proximal yet unrelated query gives $\tilde{\mathbf{h}}^{(\ell_{\min})}_e$, $\tilde{\mathbf{h}}^{(\ell_{\min})}_g$, and $\tilde{\mathbf{h}}^{(\ell_{\min})}_l$, respectively. We first impose a dual-margin loss. Let $s(\cdot,\cdot)$ denote cosine similarity. The first margin loss pulls the edit representation closer to the representation of the logically related query than to the representation of the feature-proximal yet unrelated query:
\begin{equation}
\mathcal{L}_{\text{trip1}}
=
\max\left(0, s(\tilde{\mathbf{h}}^{(\ell_{\min})}_e,\tilde{\mathbf{h}}^{(\ell_{\min})}_l)-s(\tilde{\mathbf{h}}^{(\ell_{\min})}_e,\tilde{\mathbf{h}}^{(\ell_{\min})}_g)+m\right),
\end{equation}
where $m$ is the margin. To further push representation of the feature-proximal yet unrelated query, we introduce a second margin loss:
\begin{equation}
\mathcal{L}_{\text{trip2}}
=
\max\left(0, s(\tilde{\mathbf{h}}^{(\ell_{\min})}_g,\tilde{\mathbf{h}}^{(\ell_{\min})}_l)-s(\tilde{\mathbf{h}}^{(\ell_{\min})}_e,\tilde{\mathbf{h}}^{(\ell_{\min})}_g)+m\right).
\end{equation}
Although the margin losses provide relative ordering constraints, they do not explicitly determine the absolute position of the decision boundary. Therefore, we introduce an absolute similarity regularization:
\begin{equation}
\mathcal{L}_{\text{abs}}
=
\max\left(0,1-s(\tilde{\mathbf{h}}^{(\ell_{\min})}_e,\tilde{\mathbf{h}}^{(\ell_{\min})}_g)\right)
+
\max\left(0,s(\tilde{\mathbf{h}}^{(\ell_{\min})}_e,\tilde{\mathbf{h}}^{(\ell_{\min})}_l)\right).
\end{equation}
This term anchors positive pairs toward high similarity and negative pairs toward low similarity, making the fixed routing threshold more stable during inference.

In parallel, the classification head is optimized with binary routing supervision. Let $\mathcal{P}$ denote the set of edit-related representations and $\mathcal{N}$ denote the set of unrelated representations. The binary cross-entropy loss is defined as:
\begin{equation}
\mathcal{L}_{\text{bce}}
=
\frac{1}{|\mathcal{P}|+|\mathcal{N}|}
\left(
\sum_{\tilde{\mathbf{h}}\in\mathcal{P}}\mathrm{BCE}(\boldsymbol{W}_{\text{cls}}(\tilde{\mathbf{h}}),1)
+
\sum_{\tilde{\mathbf{h}}\in\mathcal{N}}\mathrm{BCE}(\boldsymbol{W}_{\text{cls}}(\tilde{\mathbf{h}}),0)
\right),
\end{equation}
where $\boldsymbol{W}_{\text{cls}}$ denotes the classification head.
The overall disentanglement loss is:
\begin{equation} 
\mathcal{L}_{\text{dis}}
=
\lambda_1(\mathcal{L}_{\text{trip1}}+\mathcal{L}_{\text{trip2}})
+
\lambda_2\mathcal{L}_{\text{abs}}
+
\lambda_3\mathcal{L}_{\text{bce}},
\end{equation}
where $\lambda_1$, $\lambda_2$, and $\lambda_3$ balance the disentanglement, absolute similarity regularization, and binary supervision.

\section{Experiments}
\subsection{Experimental Setup}
\noindent\textbf{Base MLLMs.} To ensure a comprehensive and robust evaluation of our proposed framework, we carefully select three representative MLLMs spanning various parameter scales and architectural paradigms: BLIP2-OPT-2.7B~\citep{blip}, Gemma3-4B~\cite{gemma3}, Internvl3.5-8B~\citep{internvl3.5} as the base MLLMs.

\noindent\textbf{Baseline Methods.} We compare LDKE with representative baselines, including fine-tuning variants that update either the LLM parameters \textbf{FT(LLM)} or the visual encoder \textbf{FT(Vis)}~\citep{ft-visual}, the training-free in-context editing method \textbf{IKE}~\citep{IKE}, the meta-learning editor \textbf{MEND}~\citep{MEND}, and the multimodal knowledge editing method \textbf{MSCKE}~\citep{mscke} and \textbf{VisEdit}~\citep{visedit}.

\noindent\textbf{Datasets and Metrics.}
To comprehensively evaluate our approach, we conduct extensive experiments on FGVEdit~\citep{mscke} and VLKEB~\citep{vlkeb}. \textbf{FGVEdit} targets fine-grained multimodal editing scenarios in which edit-related queries and visually similar but unrelated queries coexist within the same image. We choose it because its fine-grained data directly supports our analysis of causal misalignment and feature entanglement. \textbf{VLKEB} provides a complementary large-scale vision-language editing benchmark with portability and sequential editing settings, allowing us to evaluate whether an edit can transfer to portability queries and remain stable across a sequence of edits. For FGVEdit, we report \textbf{Reliability}, \textbf{Generality}, text-only \textbf{Locality}, \textbf{Fine-grained Generality} (FG-Gen), and \textbf{Fine-grained Locality} (FG-Loc). For VLKEB, we report \textbf{Reliability}, \textbf{Generality}, \textbf{Locality}, \textbf{M-Generality}(M-Gen), \textbf{M-Locality}(M-Loc), \textbf{Portability}(port), and sequential-editing performance. To better characterize the fine-grained failure mode studied in this paper, we split the original FGVEdit ``Specificity'' evaluation into FG-Gen and FG-Loc according to whether the query is logically related to the edited fact. Detailed metric definitions are provided in Appendix~\ref{app:metrics}.

\subsection{Main Results}
\begin{table}[t] 
\centering
\caption{Comparison of different knowledge editing methods on FGVEdit. \textbf{Bold} and \underline{underlined} values denote the best and second-best performance, respectively.}
\label{tab:results}
\renewcommand{\arraystretch}{0.9} 
\small 
\resizebox{0.9\textwidth}{!}{
\begin{tabular}{lc ccccc} 
\toprule
\textbf{Model} & \textbf{Method} & \textbf{Reliability$\uparrow$} & \textbf{Generality$\uparrow$} & \textbf{Locality$\uparrow$} & \textbf{FG-Gen.$\uparrow$} & \textbf{FG-Loc.$\uparrow$} \\ 
\midrule
\multirow{7}{*}{BLIP2-OPT} 
    & FT(LLM)  & \textbf{100.0} & \textbf{99.96} & 76.93 & 12.56   & 35.86   \\
    & FT(Vis)  & 99.68 & \underline{99.21} & 100.0 & 8.74    & 24.38   \\
    & IKE      & \underline{99.89} & 98.02 & 48.49 & 15.95   & 24.19   \\ \cmidrule{2-7}
    & VisEdit  & 95.79 & 95.06 & 100.0 &53.86 & 47.83 \\
    & MEND     & 98.02 & 98.00 & \underline{98.43} & \underline{59.00} & \underline{71.59} \\
    & MSCKE    & 99.13 & 98.56 & 100.0 & 53.67   & 69.53  \\
    & \textbf{LDKE(Ours)} & 97.00 & 97.00 & \textbf{100.0} & \textbf{61.13} & \textbf{86.18} \\ 
\midrule
\multirow{7}{*}{Gemma3} 
    & FT(LLM)  & \textbf{100.0} & \textbf{99.00} & 70.21 & 28.00 & 23.36 \\
    & FT(Vis)  & 18.00 & 11.00 & 100.0 & 4.00  & 7.87  \\
    & IKE      & \underline{99.12} & \underline{97.88} & 50.45 & 20.21 & 30.66    \\ \cmidrule{2-7}
    & VisEdit  & 72.43 & 68.83 & 100.0 & \underline{51.05} & 43.28 \\
    & MEND     & 88.20 & 91.80 & 97.23 & 47.00 & \underline{62.81} \\
    & MSCKE    & 52.00 & 52.00 & \underline{98.20}   & 51.00   & 16.26   \\
    & \textbf{LDKE(Ours)} & 95.20 & 93.60 & \textbf{100.0} & \textbf{53.20} & \textbf{65.24} \\ 
\midrule
\multirow{7}{*}{InternVL3.5} 
    & FT(LLM)  & \textbf{100.0} & \textbf{100.0} & 56.48 & 26.00 & 17.64 \\
    & FT(Vis)  & 93.00 & 58.00 & 100.0 & 21.00 & 4.85  \\
    & IKE      & \underline{98.70} & \underline{98.10} & 58.34  & 22.48 & \underline{23.17}   \\ \cmidrule{2-7}
    & VisEdit  & 91.00 & 88.59 & 100.0 & 64.37 & 16.32 \\
    & MEND     & 94.77 & 93.09 & 96.43 & \underline{70.90} & 19.83 \\
    & MSCKE    & 68.00 & 65.00 & \underline{98.78}   & 69.00   & 12.14   \\
    & \textbf{LDKE(Ours)} & 96.00 & 91.09 & \textbf{100.0} & \textbf{73.25} & \textbf{23.54} \\ 
\bottomrule
\end{tabular}
} 
\end{table}
\paragraph{FGVEdit.}Table ~\ref{tab:results} presents the comprehensive evaluation of our proposed method against state-of-the-art baselines on FGVEdit benchmark. This consistent superiority across fundamentally different models underscores the broad applicability of our framework.
While FT(LLM) and FT(vis) achieve better performance in reliability in most models, they fall short of in FG-Gen and FG-Loc, demonstrating Fine-tune only change model's superficial output and relying solely on a distance constraint loss is fundamentally insufficient to prevent catastrophic interference on fine-grained unrelated knowledge. Though MEND achieves better results on fine-grained metrics than FT and IKE, its overall efficacy is still inherently limited. Specifically, fixing editing layers and constraining weight updates solely via distance loss prevents it from achieving optimal fine-grained disentanglement. Mscke performs well on the BLIP2-OPT model but poorly on Gemma3 and Internvl3.5, indicating very low portability to other models. In highly challenging fine-grained scenarios, the advantages of our method are further amplified. Our Fast Localization module precisely pinpoints fact-critical layers in an instance-specific manner. This dynamic adaptability enables our method to comprehensively surpass MEND in Fine-grained Generality (FG-Gen), achieving, for instance, a leading score of 73.25\% on InternVL3.5-8b compared to MEND's 71.90\%. More crucially,  thanks to disentanglement classifier, our method establishes an overwhelming advantage over MEND in Fine-grained Locality (FG-Loc), surging from 71.59\% to 88.18\% on BLIP2-OPT, perfectly realizing the bypass protection for out-of-scope fine-grained facts. VisEdit achieve better performance through Learn able modules.

\paragraph{Results on VLKEB.} Table~\ref{tab:vlkeb} presents an extensive evaluation of our method against state-of-the-art baselines on the VLKEB benchmark. Due to metric overlap with FGVEdit, we primarily focus on the Portability metric. Traditional fine-tuning methods, both FT (LLM) and FT (Vis), exhibit poor performance in Portability as they merely overfit the output probabilities without learning the underlying knowledge. In contrast, IKE consistently achieves better Portability because the edited fact is explicitly provided as contextual evidence, making it easier to trigger the model's language reasoning capabilities. Similarly, MSCKE utilizes external memory to prompt the model for one-hop reasoning, yielding stronger performance on Gemma3. Furthermore, methods like VisEdit and MEND, which leverage trained modules, also demonstrate competitive results. Distinct from these approaches, our LDKE inherently modifies the logical association of the edited knowledge through the Fast Localization module, thereby achieving superior generalization on one-hop questions.  
\begin{table}[t]
\centering
\caption{Comparison of different knowledge editing methods on VLKEB. \textbf{Bold} and \underline{underlined} values denote the best and second-best performance, respectively. Portability refers to one hop portability.}
\label{tab:vlkeb}
\renewcommand{\arraystretch}{0.9} 
\begin{tabular}{lccccccc}
\toprule
\textbf{Model} & \textbf{Method} & \textbf{Rel}$\uparrow$ & \textbf{T-Gen}$\uparrow$ & \textbf{I-Gen}$\uparrow$ & \textbf{T-Loc}$\uparrow$ & \textbf{I-Loc}$\uparrow$ & \textbf{Port}$\uparrow$ \\
\midrule
\multirow{7}{*}{Gemma3} 
 & FT(LLM)   & \textbf{100.0} & \textbf{99.67} & \textbf{100.0} & 80.30 & 33.03 & 21.34 \\
 & FT(Vis)   & 90.28 & 58.35 & 34.75 & 100.0 & 18.83 & 27.31 \\
 & IKE       & \underline{99.60} & \underline{99.56} & \underline{99.67} & 50.10 & 12.82 & \underline{43.88} \\
\cmidrule{2-8}
 & MEND      & 97.56 & 96.88 & 95.35 & \underline{99.16} & 90.35 & 37.65 \\
 & VisEdit   & 96.12 & 95.18 & 95.72 & 100.0 & 74.67 & 33.35 \\
 & MSCKE     & 95.32 & 94.85 & 95.32 & 100.0 & \textbf{99.44} & 33.33 \\
 & \textbf{LDKE(Ours)} & 96.77 & 97.50 & 96.50 & \textbf{100.0} & \underline{99.12} & \textbf{45.21} \\
\midrule
\multirow{7}{*}{InternVL3.5} 
 & FT(LLM)   & \textbf{100.0} & \textbf{98.44} & \textbf{99.30} & 87.02 & 18.26 & 1.33  \\
 & FT(Vis)   & 93.11 & 10.88 & 83.57 & 100.0 & 69.78 & 1.02  \\
 & IKE       & 95.54 & \underline{97.65} & 95.31 & 68.94 & 18.86 & \underline{13.70} \\
\cmidrule{2-8}
 & MEND      & 96.37 & 95.48 & 95.50 & 95.60 & 92.18 & 9.12  \\
 & VisEdit   & 93.83 & 91.11 & 92.20 & 99.50 & 70.68 & 10.90 \\
 & MSCKE     & 92.68 & 91.73 & 94.37 & \underline{99.95} & \underline{98.87} & 1.40  \\
 & \textbf{LDKE(Ours)} & \underline{97.78} & 97.00 & \underline{98.62} & \textbf{100.0} & \textbf{99.78} & \textbf{18.62} \\
\bottomrule
\end{tabular}
\end{table}
\subsection{Sequential Editing}
To evaluate sequential editing capabilities, we conduct experiments involving 10 and 100 continuous edits using MSCKE, our LDKE, and LiveEdit, a specialized sequential editing framework for MLLMs.  Table ~\ref{fig:sequential_editing} records results on reliability, image generality, image locality and more results refer to Appendix~\ref{sequential}. LiveEdit maintains highly consistent performance across all metrics even after 100 sequential edits. In contrast, both MSCKE and LDKE exhibit varying degrees of degradation. Notably, the performance of LDKE drops to near zero across Reliability, Generality, I-Generality, and Portability after merely 10 edits. We attribute this severe degradation to LDKE's reliance on a MEND-style updater for weight modifications. During sequential editing, the model undergoes continuous parameter shifts. Consequently, the MEND-style editor, which is optimized based on the parameter distribution of the original model, progressively loses its efficacy as the model deviates further from its initial state, ultimately leading to a catastrophic performance drop.
\begin{figure}[t]
    \centering
    
    \includegraphics[width=\linewidth,draft=false]{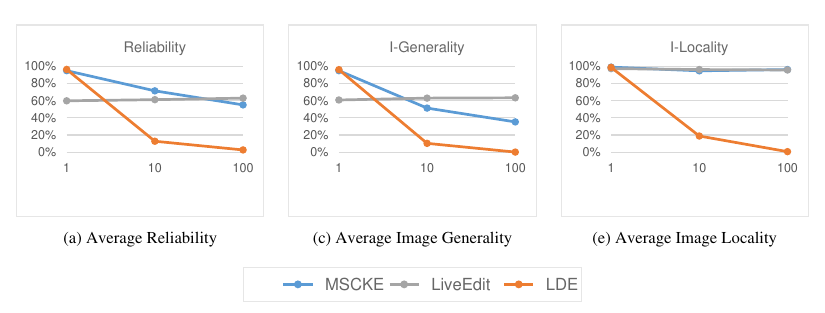}
    \caption{Average results in sequential editing.}
    \label{fig:sequential_editing}
\end{figure}

\subsection{Effect of the Disentanglement Classifier}

\begin{table}[htbp]
\centering
\caption{Ablation study on the representational disentanglement of different classifiers on BLIP2-OPT. Values denote the average cosine similarity between the edit sample and various evaluation sample categories, expressed as percentages (\%).}
\label{tab:ablation_classifier}
\resizebox{0.8\textwidth}{!}{
\begin{tabular}{l c c c c c}
\toprule
\textbf{Classifier Variants} & \textbf{Edit} $\uparrow$ & \textbf{Gen} $\uparrow$ & \textbf{Loc} $\downarrow$ & \textbf{FG-Gen} $\uparrow$ & \textbf{FG-Loc} $\downarrow$ \\ 
\midrule
Last Token Classifier        & 99.20 & 94.70 & 17.30 & 75.40 & 67.70 \\ 
Disentanglement Classifier   & 98.10 & 93.16 & 6.70  & 69.66 & 30.03 \\ 
\bottomrule
\end{tabular}
}
\end{table}

Table \ref{tab:ablation_classifier} presents the average cosine similarity between the representation of the edit sample and various evaluation sample categories. When directly utilizing the raw hidden states of the final prompt token (Last Token Classifier), the model suffers from severe feature entanglement. Consequently, the representations of in-scope relevant samples (FG-Gen) and out-of-scope irrelevant samples (FG-Loc) within the same visual context exhibit dangerously close similarity scores (75.40 vs. 67.70). This narrow margin renders the classifier unable to establish a robust decision boundary. In contrast, our proposed Disentanglement Classifier successfully decouples these representations. While maintaining sufficiently high similarity for relevant facts (FG-Gen at 69.66), it suppresses the similarity score for fine-grained irrelevant knowledge (dropping FG-Loc to 30.03). This substantial margin disentangles the visual scope, enabling the classifier to highly accurately distinguish  fine-grained inputs.

\subsection{Effect of Different Edited Layers}

\begin{table}[htbp]
\centering
\caption{Ablation study on layer selection strategies within Layer-Specific Weight Editor on Gemma3.}
\label{tab:edited_layers}
\renewcommand{\arraystretch}{1.2}
\resizebox{0.9\textwidth}{!}{
\begin{tabular}{l ccccc}
\toprule
\textbf{Layer Selection Strategy} & \textbf{Reliability$\uparrow$} & \textbf{Generality$\uparrow$} & \textbf{Locality$\uparrow$} & \textbf{FG-Gen.$\uparrow$} & \textbf{FG-Loc.$\uparrow$} \\ 
\midrule
Last three layers                 & 94.77 & 93.09 & 96.43 & 70.90 & 19.83 \\ 
Layers selected by causal tracing & 95.22 & 93.33 & 96.78 & 75.01 & 17.43 \\ 
Fast Localization (Ours)          & 96.00 & 91.09 & 95.90 & 76.25 & 19.66 \\ 
\bottomrule
\end{tabular}
}
\end{table}

We present the results of different layer selection strategies in Table~\ref{tab:edited_layers}. When utilizing meta-learning networks, defaulting to the last three layers yields the worst performance in \textbf{FG-Gen.}. This is because the final layers of the model are primarily dedicated to input integration; editing at this stage can only fit superficial semantic mappings. Employing multimodal causal tracing to locate and edit layers with the highest contributions achieves better performance than the static baseline. A substantially more effective approach is to dynamically select the target editing layers based on each individual input instance. By introducing our Fast Localization module, we bypass this bottleneck and make instance-level dynamic selection computationally feasible, ultimately allowing our method to achieve the best  generalization results.

\section{Conclusion}
This paper identified the generalization-localization challenge in multimodal knowledge editing, driven by causal misalignment and feature entanglement. To mitigate these issues, We presents the Localized and Disentangled Knowledge Editing (LDKE) framework to address causal misalignment and feature entanglement. By a Fast Localization module and a Disentanglement Classifier, LDKE successfully overcomes these challenges. Extensive evaluations on the multiple benchmarks across various MLLMs demonstrate that our method significantly outperforms existing baselines.


\bibliographystyle{IEEEtran}  
\bibliography{main}           
\newpage
\appendix
\section{Limitations} 
\label{limit}
A primary limitation of LDKE lies in its suboptimal performance during sequential editing. This vulnerability stems from our adoption of a MEND-style weight editor, which inherently struggles with the parameter shift problem. Because the MEND-style updater is trained on the parameter space of the original model, continuous modifications during sequential editing cause the model's weights to drift. Consequently, the updater gradually becomes misaligned and loses its efficacy. Furthermore, methods that directly modify MLLM parameters at scale frequently suffer from representation interference; subsequent edits tend to overwrite or disrupt previously injected knowledge, leading to severe performance degradation. In future work, we plan to extend LDKE to the sequential editing paradigm, aiming to mitigate parameter drift and maintain consistently stable performance across continuous edits. 

\section{Broader Impacts}
\label{impacts}
This work aims to advance reliable multimodal knowledge editing by improving the precision and controllability of factual updates in MLLMs. LDKE is designed to correct target knowledge while reducing unintended interference with fine-grained, visually related but logically unrelated facts. This capability can support safer model maintenance in applications such as factual correction, continual model updating, hallucination mitigation, and domain-specific AI systems that require timely knowledge revision without full retraining. 

At the same time, knowledge editing techniques can be misused to inject biased, misleading, or unauthorized information into deployed models. We therefore emphasize the need for responsible deployment practices, including edit provenance, validation on both related and unrelated queries, access control, and auditing mechanisms that make model updates traceable and reversible.

\section{Experiement Details}
\label{app:metrics}
\subsection{Evaluation Metric Details}
\paragraph{FGVEdit metrics.}
For an edit instance $(i_e,t_e,y_e)$, \textbf{Reliability} measures whether the edited model predicts the target answer on the original edit query:
\begin{equation}
\text{Rel} =
\mathbb{E}_{(i_e,t_e,y_e)\in\mathcal{D}_{\text{edit}}}
\left[\mathbb{I}\left(f_{\theta_e}(i_e,t_e)=y_e\right)\right].
\end{equation}
\textbf{Generality} evaluates whether the edited knowledge transfers to semantically equivalent prompts:
\begin{equation}
\text{Gen} =
\mathbb{E}_{(i_e,t_{\text{gen}},y_e)\in\mathcal{D}_{\text{gen}}}
\left[\mathbb{I}\left(f_{\theta_e}(i_e,t_{\text{gen}})=y_e\right)\right].
\end{equation}
\textbf{Locality} measures whether the edited model preserves the original model's prediction on unrelated text-only queries:
\begin{equation}
\text{Loc} =
\mathbb{E}_{t_{\text{loc}}\in\mathcal{D}_{\text{loc}}}
\left[\mathbb{I}\left(f_{\theta_e}(t_{\text{loc}})=f_\theta(t_{\text{loc}})\right)\right].
\end{equation}
We further split the original FGVEdit specificity evaluation into two fine-grained metrics. \textbf{FG-Gen} evaluates in-scope visual queries that remain logically related to the edit:
\begin{equation}
\text{FG-Gen} =
\mathbb{E}_{(i_e,t_{\text{in}},y_{\text{in}})\in\mathcal{D}_{\text{in}}}
\left[\mathbb{I}\left(f_{\theta_e}(i_e,t_{\text{in}})=y_{\text{in}}\right)\right].
\end{equation}
\textbf{FG-Loc} evaluates out-of-scope visual queries from the same image whose predictions should remain unchanged:
\begin{equation}
\text{FG-Loc} =
\mathbb{E}_{(i_e,t_{\text{out}})\in\mathcal{D}_{\text{out}}}
\left[\mathbb{I}\left(f_{\theta_e}(i_e,t_{\text{out}})=f_\theta(i_e,t_{\text{out}})\right)\right].
\end{equation}

\paragraph{VLKEB metrics.}
VLKEB uses \textbf{Reliability}, \textbf{Generality}, and \textbf{Locality} to evaluate edit success, paraphrase robustness, and preservation of unrelated knowledge. It additionally includes \textbf{Portability}, which measures whether the edited fact transfers to related portability queries:
\begin{equation}
\text{Port} =
\mathbb{E}_{(q_{\text{port}},y_{\text{port}})\in\mathcal{D}_{\text{port}}}
\left[\mathbb{I}\left(f_{\theta_e}(q_{\text{port}})=y_{\text{port}}\right)\right].
\end{equation}
For sequential editing, we apply a sequence of edits and report the same evaluation dimensions after the final edited model, measuring whether performance remains stable under accumulated updates.

\subsection{Hyperparameters}
\label{hyper}

\textbf{IKE} Use the sentence-transformers model all-MiniLM-L6-v2 to embed texts and retrieve similar edits in training set. The number of demonstrations is set to 32 for all models.

\begin{table}[H] 
\centering
\caption{Hyperparameters for different knowledge editing methods.}
\label{tab:hyperparameters}
\begin{tabular}{llclc}
\toprule
\textbf{Method} & \textbf{Model} & \textbf{Steps / MaxIter} & \textbf{Edit Layer} & \textbf{LR} \\
\midrule
\multirow{3}{*}{FT-LLM} 
& BLIP2-OPT & 1 & 31$^{\text{st}}$ layer of Transformer Module & 1e-5 \\
& Gemma3 & 1 & 33$^{\text{rd}}$ layer of Language Model & 1e-5 \\
& InternVL3.5 & 10 & layer 33--35 of Language Model & 1e-4 \\
\midrule
\multirow{3}{*}{FT-Vis} 
& BLIP2-OPT & 15 & Qformer & 1e-5 \\
& Gemma3 & 15 & 26$^{\text{th}}$ layer of Vision Encoder & 1e-5 \\
& InternVL3.5 & 10 & MLP projector (mlp1) & 1e-4 \\
\midrule
\multirow{3}{*}{MEND} 
& BLIP2-OPT & 30000 & layer 29, 30, 31 of Transformer Module & 1e-6 \\
& Gemma3 & 30000 & layer 31, 32, 33 of Language Model & 1e-6 \\
& InternVL3.5 & 20000 & layer 33, 34, 35 of Language Model & 1e-6 \\
\midrule
\multirow{3}{*}{MSCKE} 
& BLIP2-OPT & 30000 & layer 20--29 of Transformer Module & 1e-6 \\
& Gemma3 & 30000 & layer 20--29 of Language Model & 1e-6 \\
& InternVL3.5 & 20000 & layer 33, 34, 35 of Language Model & 1e-5 \\
\midrule
\multirow{3}{*}{LDKE} 
& BLIP2-OPT & 30000 & determined by Fast Localization Module & 1e-6 \\
& Gemma3 & 30000 & determined by Fast Localization Module & 1e-6 \\
& InternVL3.5 & 20000 & determined by Fast Localization Module & 1e-6 \\
\midrule
\multirow{3}{*}{VisEdit} 
& BLIP2-OPT & 30000 & 19$^{\text{th}}$ layer of Transformer Module; Qformer & 1e-4 \\
& Gemma3 & 30000 & 26$^{\text{th}}$ layer of LM; multi-modal projector & 1e-4 \\
& InternVL3.5 & 20000 & 24$^{\text{th}}$ layer of LM; MLP projector (mlp1) & 1e-4 \\
\bottomrule
\multicolumn{5}{l}{\footnotesize *All methods use the Adam optimizer.} \\
\end{tabular}
\end{table}

\subsection{Experiement settings}
\label{exper settings}
All experiments were implemented using the PyTorch framework and conducted on a server equipped with eight NVIDIA A100-SXM4 80GB GPUs and Intel Xeon Platinum 8368 CPU.

\section{Additional Experiements}
\subsection{Analysis of Time Cost Across Different Localization Modules}

The Table~\ref{tab:time_cost} presents a comparative analysis of the computational overhead introduced by different localization modules relative to the standard inference latency. The computational cost of the causal tracing module is significant, requiring approximately 571,250 ms per data instance. In contrast, the Fast Localization module is highly efficient, requiring only 4.66 ms per sample to identify layers critical to factual association. Compared to the total inference latency of 133 ms, this computational overhead is negligible. Therefore, Fast Localization can dynamically select edited layers for each instance, leading to better performance than alternative localization modules.

\begin{table}[htbp]
\centering
\caption{Comparison of time cost across different localization modules on BLIP2-OPT. Our Fast Localization module significantly reduces the overhead compared to the causal tracing module.}
\label{tab:time_cost}
\renewcommand{\arraystretch}{1.3} 
\setlength{\tabcolsep}{8pt}      
\begin{tabular}{lccc} 
\toprule
Module & Standard Inference & Causal Tracing & \textbf{Fast Localization} \\ 
\midrule
Time (ms) & 78.49 & 391751.28 & \textbf{4.66} \\ 
\bottomrule
\end{tabular}
\end{table}

\subsection{Sequential Experiements}
\label{sequential}
Fig.~\ref{fig:sequential_editing} shows more results on sequential editing.

\begin{figure}[h]
    \centering
    
    \includegraphics[width=\linewidth,draft=false]{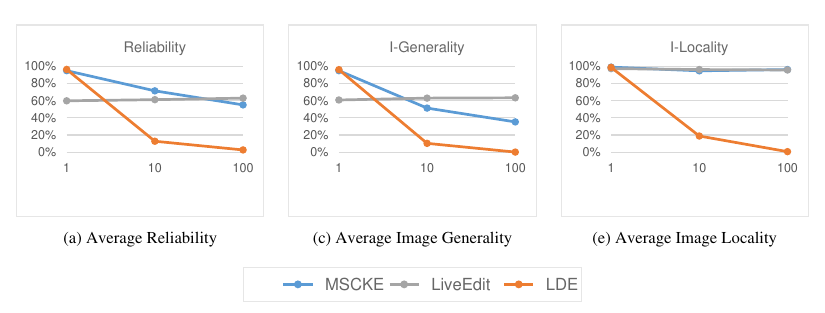}
    \caption{Average results in sequential editing.}
    \label{fig:sequential_editing}
\end{figure}



\clearpage
\section*{NeurIPS Paper Checklist}

\begin{enumerate}

\item {\bf Claims}
    \item[] Question: Do the main claims made in the abstract and introduction accurately reflect the paper's contributions and scope?
    \item[] Answer: \answerYes{} 
    \item[] Justification: The abstract and introduction clearly summarize the main contributions and scope of the paper.

\item {\bf Limitations}
    \item[] Question: Does the paper discuss the limitations of the work performed by the authors?
    \item[] Answer: \answerYes{} 
    \item[] Justification: The limitations of our work are explicitly discussed in a dedicated Appendix~\ref{limit}.

\item {\bf Theory assumptions and proofs}
    \item[] Question: For each theoretical result, does the paper provide the full set of assumptions and a complete (and correct) proof?
    \item[] Answer: \answerNA{} 
    \item[] Justification: The paper does not include theoretical results

    \item {\bf Experimental result reproducibility}
    \item[] Question: Does the paper fully disclose all the information needed to reproduce the main experimental results of the paper to the extent that it affects the main claims and/or conclusions of the paper (regardless of whether the code and data are provided or not)?
    \item[] Answer: \answerYes{} 
    \item[] Justification: Appendix~\ref{hyper} reports the base models and experiment parameters needed to reproduce the experiment.

\item {\bf Open access to data and code}
    \item[] Question: Does the paper provide open access to the data and code, with sufficient instructions to faithfully reproduce the main experimental results, as described in supplemental material?
    \item[] Answer: \answerNo{} 
    \item[] Justification: The code repository is undergoing further refinement.

\item {\bf Experimental setting/details}
    \item[] Question: Does the paper specify all the training and test details (e.g., data splits, hyperparameters, how they were chosen, type of optimizer) necessary to understand the results?
    \item[] Answer: \answerYes{} 
    \item[] Justification: Please refer to Appendix~\ref{hyper} for information about Experimental setting/details.

\item {\bf Experiment statistical significance}
    \item[] Question: Does the paper report error bars suitably and correctly defined or other appropriate information about the statistical significance of the experiments?
    \item[] Answer: \answerNo{} 
    \item[] Justification: The paper does not include error bars or statistical significance tests.

\item {\bf Experiments compute resources}
    \item[] Question: For each experiment, does the paper provide sufficient information on the computer resources (type of compute workers, memory, time of execution) needed to reproduce the experiments?
    \item[] Answer: \answerYes{} 
    \item[] Justification: pleaser refer to Appendix~\ref{exper settings} for more information about computer resources.
    
\item {\bf Code of ethics}
    \item[] Question: Does the research conducted in the paper conform, in every respect, with the NeurIPS Code of Ethics \url{https://neurips.cc/public/EthicsGuidelines}?
    \item[] Answer: \answerYes{} 
    \item[] Justification:  Our work conforms to the NeurIPS Code of Ethics.

\item {\bf Broader impacts}
    \item[] Question: Does the paper discuss both potential positive societal impacts and negative societal impacts of the work performed?
    \item[] Answer: \answerYes{} 
    \item[] Justification: We discuss both the potential positive and negative societal impacts of our work in Appendix~\ref{impacts}.
    
\item {\bf Safeguards}
    \item[] Question: Does the paper describe safeguards that have been put in place for responsible release of data or models that have a high risk for misuse (e.g., pre-trained language models, image generators, or scraped datasets)?
    \item[] Answer: \answerNA{} 
    \item[] Justification: Our work does not involve the release of any models or datasets with high risk of misuse.

\item {\bf Licenses for existing assets}
    \item[] Question: Are the creators or original owners of assets (e.g., code, data, models), used in the paper, properly credited and are the license and terms of use explicitly mentioned and properly respected?
    \item[] Answer: \answerYes{} 
    \item[] Justification: We use existing open-source datasets and pretrained models, all of which are properly cited in the main text. Each open-source dataset and model is accompanied by license and usage terms, where available.

\item {\bf New assets}
    \item[] Question: Are new assets introduced in the paper well documented and is the documentation provided alongside the assets?
    \item[] Answer: \answerNA{} 
    \item[] Justification: The paper does not release new assets.

\item {\bf Crowdsourcing and research with human subjects}
    \item[] Question: For crowdsourcing experiments and research with human subjects, does the paper include the full text of instructions given to participants and screenshots, if applicable, as well as details about compensation (if any)? 
    \item[] Answer: \answerNA{} 
    \item[] Justification: The paper does not involve crowdsourcing nor research
with human subjects.

\item {\bf Institutional review board (IRB) approvals or equivalent for research with human subjects}
    \item[] Question: Does the paper describe potential risks incurred by study participants, whether such risks were disclosed to the subjects, and whether Institutional Review Board (IRB) approvals (or an equivalent approval/review based on the requirements of your country or institution) were obtained?
    \item[] Answer: \answerNA{} 
    \item[] Justification: The paper does not involve crowdsourcing nor research
with human subjects.

\item {\bf Declaration of LLM usage}
    \item[] Question: Does the paper describe the usage of LLMs if it is an important, original, or non-standard component of the core methods in this research? Note that if the LLM is used only for writing, editing, or formatting purposes and does \emph{not} impact the core methodology, scientific rigor, or originality of the research, declaration is not required.
    \item[] Answer: \answerYes{} 
    \item[] Justification: We position large language models as the primary subject of our research.

\end{enumerate}

\end{document}